\documentclass[%
 reprint,
superscriptaddress,
 amsmath,amssymb,
 aps,
 prx,
]{revtex4-2}
\usepackage{appendix}
\usepackage{amsthm}

\usepackage{booktabs}
\usepackage{mathtools}
\usepackage{graphicx}
\graphicspath{{./}}
\usepackage{dcolumn}
\usepackage{bm}

\usepackage[
  colorlinks=true,
  linkcolor=blue,
  citecolor=blue,
  urlcolor=blue
]{hyperref}

\newcommand{\fc}{f_{\mathrm{c}}}

\newcommand{\chim}{\chi_{\mathrm{max}}}

\begin{document}

\preprint{APS/123-QED}

\title{Grokking as a Falsifiable Finite-Size Transition}

\author{Yuda Bi}
\email{ybi3@gsu.edu}
\affiliation{%
Tri-Institutional Center for Translational Research in Neuroimaging and Data Science (TReNDS Center),\\
Atlanta, Georgia 30303, USA
}%

\author{Chenyu Zhang}
\affiliation{%
Beijing Institute of Technology, Beijing 100081, China
}%

\author{Qiheng Wang}
\affiliation{%
Beijing Institute of Technology, Beijing 100081, China
}%

\author{Vince D Calhoun}
\affiliation{%
Georgia Institute of Technology, Atlanta, Georgia 30332, USA
}%
\affiliation{%
Tri-Institutional Center for Translational Research in Neuroimaging and Data Science (TReNDS Center),\\
Atlanta, Georgia 30303, USA
}%

\begin{abstract}
Grokking---the delayed onset of generalization after early memorization---is often described with phase-transition language, but that claim has lacked falsifiable finite-size inputs.
Here we supply those inputs by treating the group order $p$ of $\mathbb{Z}_p$ as an admissible extensive variable and a held-out spectral head--tail contrast as a representation-level order parameter, then apply a condensed-matter-style diagnostic chain to coarse-grid sweeps and a dense near-critical addition audit.
Binder-like crossings reveal a shared finite-size boundary, and susceptibility comparison strongly disfavors a smooth-crossover interpretation ($\Delta\mathrm{AIC}=16.8$ in the near-critical audit).
Phase-transition language in grokking can therefore be tested as a quantitative finite-size claim rather than invoked as analogy alone, although the transition order remains unresolved at present.
\end{abstract}


\maketitle

\section{Introduction}
\label{sec:introduction}

Neural networks on modular arithmetic tasks often memorize quickly and then generalize only after long optimization, a delayed phenomenon now widely called \emph{grokking}~\cite{power2022grokking,liu2023omnigrok,nanda2023progress,lee2024grokfast,humayun2024deep}.
The effect is sharp enough that transition language is now routine: grokking is described as a ``phase transition'' in which the network reorganizes from a memorizing to a generalizing regime.
But steep training curves observed at a single system size do not by themselves constitute a falsifiable finite-size claim.
Without a legitimate size variable and an admissible order parameter, the discussion remains descriptive rather than diagnostic.

In equilibrium statistical mechanics, where the same tension between sharp features and genuine singularities arises in any finite system, finite-size scaling (FSS) provides precisely the sequential diagnostic protocol needed to resolve such ambiguity~\cite{fisher1972scaling,privman1990finite,goldenfeld1992lectures,stanley1971introduction}.
The logic is layered: one first identifies a size variable and an order parameter, then checks whether Binder cumulant curves for different sizes cross at a common control-parameter value~\cite{binder1981finite}, then tests whether fluctuation peaks grow as a power law rather than saturating, and only then asks about transition order.
Each layer has a defined failure mode, and the chain can be rejected at any step.
This sequential falsifiability is what makes FSS a diagnostic tool rather than a fitting exercise---and what distinguishes it from the common practice in machine-learning studies of fitting a sigmoidal curve to a single training run and declaring a transition---a procedure that the FSS literature would regard as necessary but far from sufficient~\cite{fisher1972scaling,binder1981finite}.

Applying this protocol to learning systems, however, requires two inputs that are not automatic: an extensive size variable and a representation-level order parameter.
The broader statistical-mechanics-of-learning tradition supplies a rich vocabulary of order parameters and scaling relations~\cite{engel2001statistical,bahri2020statistical,canatar2023statistical,ziyin2023phase}.
Representation-geometry perspectives likewise suggest that structured low-dimensional observables should exist in trained networks~\cite{tishby2015deep,saxe2019information,stringer2019high,ansuini2019intrinsic,amari2016information,fefferman2016testing,gallego2017neural,gao2015simplicity}, yet neither tradition has assembled these ingredients into a falsifiable finite-size chain for grokking.
The usual machine-learning scaling variables---width, depth, and parameter count---move across model classes rather than along a fixed task family, so they do not supply the controlled, single-family size variation that FSS requires~\cite{kaplan2020scaling}.
Meanwhile, readout-level quantities such as training loss and test accuracy do not probe the internal geometry of representations and therefore cannot serve as order parameters in the statistical-mechanics sense.
The first obstacle is therefore upstream of any exponent fit: specifying \emph{what to scale} and \emph{what to measure}.

Recent work frames grokking within incompatible physical pictures---mean-field first-order~\cite{rubin2024grokking}, critical~\cite{zunkovic2024grokking,liu2022effective,clauw2024information,demoss2025complexity}, mechanistic~\cite{nanda2023progress,merrill2023tale,lyu2024dichotomy,chughtai2023toy}, glass~\cite{zhang2025glass}---but these do not share a common finite-width diagnostic chain, so their claims cannot yet be compared on common ground.
The point, then, is not merely to import Binder curves into machine learning, but to ask whether delayed generalization can be organized by the same finite-size logic that distinguishes crossover from genuine transition in many-body systems.
That requires specifying what is being enlarged, what is ordering, and what observations would count as failure of the transition picture.
What is missing is therefore not another interpretation but the standard of individually falsifiable links familiar from condensed matter~\cite{sethna2021statistical,landau2013statistical,bahri2024explaining}.

Here we supply the two missing inputs for the canonical modular-addition benchmark.
We treat the group order $p$ of the cyclic group $\mathbb{Z}_p$ as the size variable and a held-out spectral head--tail contrast (HTC) as the order parameter, then apply Binder crossings, susceptibility comparison, and a near-critical audit as a first-layer finite-size diagnostic protocol.
The resulting evidence supports transition-like finite-size organization and strongly disfavors a smooth-crossover interpretation, while leaving transition order, asymptotic exponents, and universality outside the main claim.
Section~\ref{sec:order-param} specifies the two identifications; Section~\ref{sec:evidence} presents the diagnostic chain; Section~\ref{sec:discussion} closes with scope and implications.

\section{Setup: Two Key Identifications}
\label{sec:order-param}

\subsection{System size: why the group order $p$}
\label{sec:system-size}

\begin{figure*}[t]
\centering
\includegraphics[width=\textwidth]{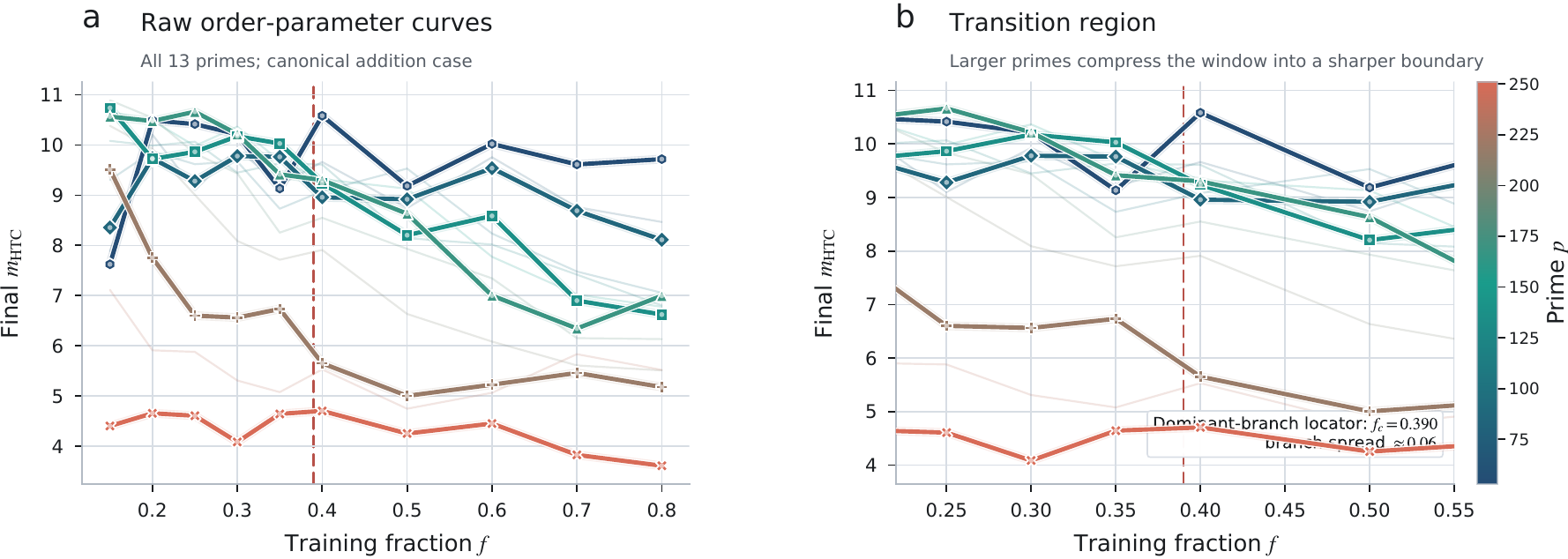}
\caption{Raw spectral order-parameter curves for the canonical addition task. Panel A shows $m_{\mathrm{HTC}}$ versus training fraction $f$ for all $13$ coarse-grid primes; Panel B zooms the transition region around the shared crossing in this first diagnostic step. The ordering structure sharpens monotonically with $p$.}
\label{fig:raw-htc}
\end{figure*}

Finite-size scaling requires an extensive variable that enlarges one controlled family rather than switching among qualitatively different systems~\cite{fisher1972scaling,privman1990finite,cardy1988finite}.
The group order $p$ is chosen because it indexes the algebraic task family itself while leaving architecture, optimizer, regularization, and logging protocol fixed.
Using $p^2$ would merely count ordered examples; it would not identify the family-defining control.
For prime groups, $\mathbb{Z}_p$ remains cyclic for every $p$, so varying $p$ enlarges one homogeneous algebraic class rather than introducing subgroup structure from composite moduli.
In this sense $p$ is not a literal spatial length, but it is the admissible extensive variable the network must resolve: the number of distinct group elements sets both the classification space and the combinatorial resolution demanded of the learned representation.
This is precisely the quantity that grows while the learning system itself is held fixed.
We do not claim that $p$ is the only conceivable scaling variable, only that it preserves the task family while avoiding model-class changes that would arise in width-, depth-, or parameter-count sweeps.
The present claim is therefore finite-size diagnostics on a fixed algebraic task family, not a literal thermodynamic-limit theorem for arbitrary architectures.
An important caveat is that the architecture is held fixed at $d_{\mathrm{model}}=128$: if $p$ were taken far beyond the present range, one would expect task extent and model capacity to compete, so the relevant bottleneck could shift.
Empirically, the explored prime range is broad enough that sharpening and near-common Binder crossings organize coherently when plotted against $p$, which is the consistency check one would want from an admissible finite-size control (Figs.~\ref{fig:raw-htc} and~\ref{fig:binder-continuity}).

At fixed $(p,f,\mathrm{op})$, the train/eval/probe partition is generated once from a task-level data seed and shared across all training seeds, so Binder and susceptibility fluctuations probe initialization and optimization variability at fixed task instance.

\subsection{Order parameter: why spectral head-tail contrast}
\label{sec:order-param-def}

The spectral head-tail contrast is defined as
\begin{equation}
\begin{aligned}
m_{\mathrm{HTC}}(t)
&= \log\!\left(
\frac{\sum_{j=1}^{5} p_j(t)}
{\sum_{j=6}^{d} p_j(t)}
\right),
\end{aligned}
\label{eq:htc-def}
\end{equation}
where $p_j(t) = \lambda_j(t)/\sum_k \lambda_k(t)$ are normalized eigenvalues of the probe-set covariance matrix $C_t$ of hidden representations $z_t \in \mathbb{R}^d$.
In implementation, a tiny stabilizer $\varepsilon = 10^{-10}$ is added to numerator and denominator for numerical safety at extreme spectral concentration.
A representation-level observable is needed because grokking is not merely a late change in readout accuracy; it is accompanied by a reorganization of internal geometry.
Readout-level quantities such as training loss and test accuracy are therefore useful diagnostics but not natural order parameters for the statistical-mechanics question posed here~\cite{martin2021implicit,churchland2010stimulus}.
Prior mechanistic work on modular arithmetic suggests that grokking is accompanied by reorganization of that internal geometry---from diffuse or memorization-specific activations to structured Fourier representations---not merely a late change in readout accuracy~\cite{nanda2023progress,chughtai2023toy,papyan2020prevalence}.
At each logged checkpoint, the model is evaluated on the held-out probe subset, the mean-pooled penultimate representations are collected and centered, and their covariance matrix is diagonalized; the normalized eigenvalues then define the spectral masses entering the HTC definition.
When the spectrum is diffuse, head and tail masses are comparable and $m_{\mathrm{HTC}}$ is small; when a few modes separate from the bulk, $m_{\mathrm{HTC}}$ grows.
HTC is therefore not introduced as a classifier of phases, but as a compact scalar probe of whether spectral mass remains distributed across the covariance bulk or condenses into a small leading sector.
The log-ratio form keeps that probe unbounded, so strongly ordered runs are not artificially compressed into a saturating score, in the spirit of low-dimensional observables that summarize collective reorganization in high-dimensional systems~\cite{goldenfeld1992lectures,machta2013parameter,ott2008low}.
Fourier-mode amplitudes are mechanistically informative, but they presuppose a task-specific basis and a more explicit circuit story.
HTC instead compresses the same representational reorganization into a held-out covariance-level scalar that remains basis-agnostic and directly compatible with finite-size diagnostics.
The choice $k=5$ is operational rather than metaphysical: it separates a small leading sector from the covariance bulk in the present representation dimension, while Appendix Fig.~\ref{fig:app-k-robustness} shows that $k=3$ preserves the same diagnostic verdict and $k=10$ weakens the separation.
Screening across alternative observables (Appendix Fig.~\ref{fig:screening}) is therefore used as validation, not as tuning to maximize crossings.

\subsection{Experimental protocol}
\label{sec:exp-design}

All results use one fixed Transformer family: $d_{\mathrm{model}}=128$, two encoder layers, four attention heads, feedforward dimension $512$, pre-LayerNorm architecture, no dropout, and mean pooling over the sequence dimension as the readout.
Optimization is likewise fixed throughout: AdamW with learning rate $5\times10^{-4}$, weight decay $1.0$, batch size $512$, and observables logged every $25$ steps up to $40{,}000$ training steps.
The full dataset for each modular task consists of all $p^2$ ordered pairs in $\mathbb{Z}_p \times \mathbb{Z}_p$; at fixed $(p,f)$, the first $\lfloor f p^2 \rfloor$ examples form the training set, a disjoint probe subset (capped at $7{,}600$ examples) is reserved for representation geometry, and the remainder serves as the held-out evaluation set.
This split is generated once per condition and shared across all seeds, so the seed ensemble probes initialization and optimization variability at fixed task instance rather than repeated repartitioning.
The coarse sweep is the broad evidence base: $13$ primes, $10$ training fractions, and $50$ seeds per condition on each operation, supplemented by a contextual phase map and Fourier probe.
The near-critical addition sweep is an audit rather than a discovery run: it re-queries the same diagnostic chain on $6$ larger primes and $11$ closely spaced fractions, with the same model, optimizer, split logic, and observable definition.
Addition is emphasized not because subtraction or multiplication fail, but because it is the cleanest algebraic benchmark in which the finite-size inputs can first be tested without extra complication; the other operations remain supporting checks (Appendix~\ref{app:experimental-details}).
All main-text summaries are stationary-window measurements of the final regime rather than arbitrary snapshots from the training trajectory.

\section{Results: Transition Diagnostics}
\label{sec:evidence}

The coarse evidence base for the canonical addition task consists of $13$ primes, $10$ training fractions, and $50$ seeds per condition, while the near-critical follow-up adds $6$ larger primes and $11$ closely spaced fractions under the same protocol.
The first sweep supplies broad finite-size coverage; the second re-queries the same diagnostics in the transition region.
The diagnostic chain proceeds in order of increasing stringency: raw sharpening, crossing consistency, crossover rejection, and transition-order assessment.

\subsection{Raw sharpening}
\label{sec:evidence-raw}

The simplest finite-size observation is that the final spectral order parameter sharpens systematically with system size.
Figure~\ref{fig:raw-htc} shows $m_{\mathrm{HTC}}(f,p)$ for the canonical addition case across all $13$ primes.
At small $p$, the transition from low to high spectral concentration is broad.
As $p$ increases, the same transition sharpens and localizes near a common fraction, which is the expected finite-size precursor of a sharp phase boundary.
The largest primes do not merely shift upward; they compress the transition window itself, producing a visibly steeper turnover in the same low-$f$ neighborhood.
No rescaling or fitting is applied: the size-resolved ordering is evident in the raw data.

\begin{figure*}[t]
\centering
\includegraphics[width=0.9\textwidth]{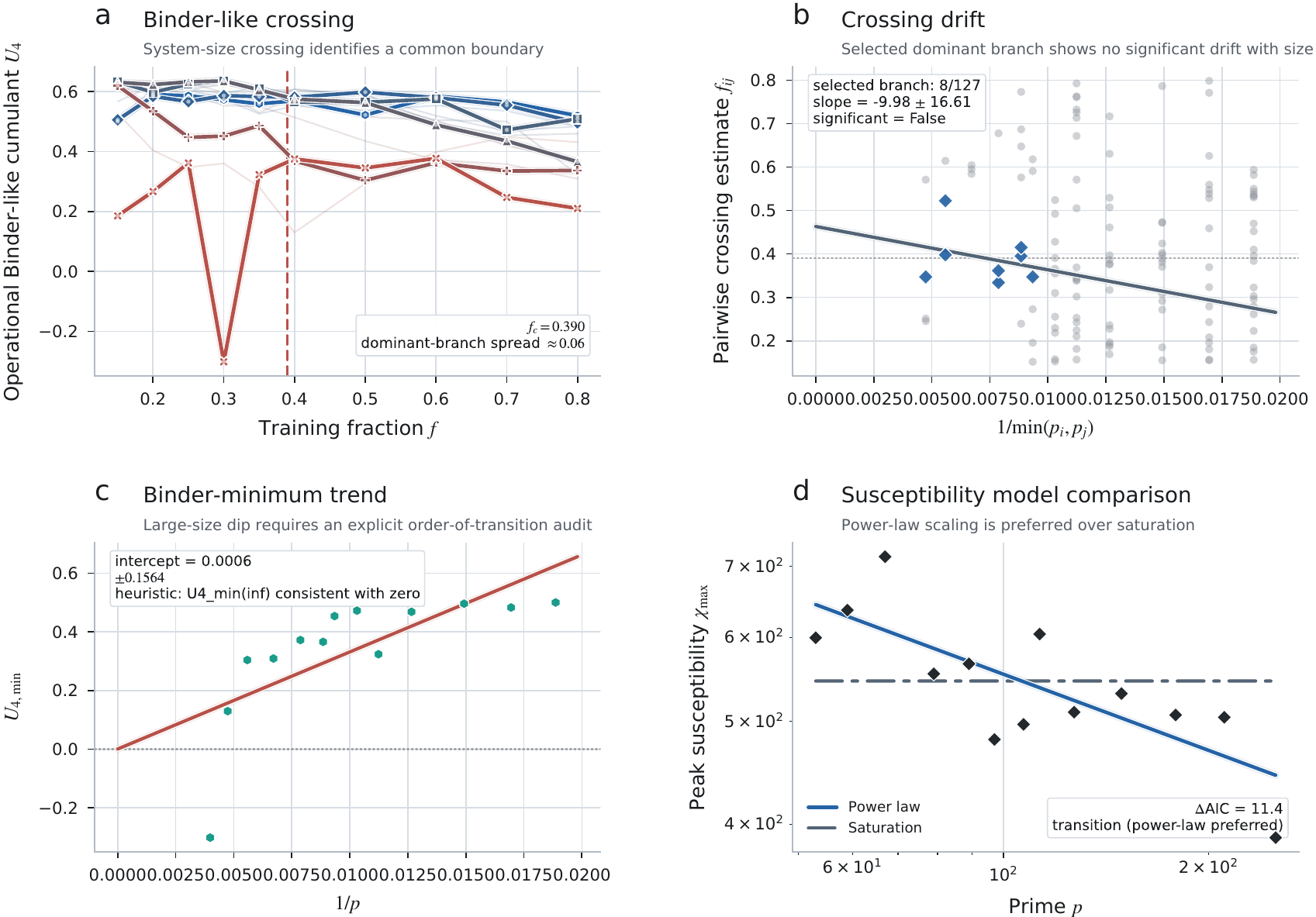}
\caption{Binder-based diagnostics for the canonical addition task. Panel A: Binder-like cumulant curves cross near a common fraction. Panel B: all pairwise crossing estimates are shown, with the selected dominant branch highlighted for the drift fit; that branch shows no significant linear drift versus inverse size. Panel C: the coarse-grid Binder minimum is continuity-leaning but method-dependent, motivating the near-critical stress test in Fig.~\ref{fig:nearcrit}A. Panel D: susceptibility model comparison favors power-law scaling over the minimal saturating null, disfavoring a smooth-crossover interpretation.}
\label{fig:binder-continuity}
\end{figure*}

\subsection{Binder crossing}
\label{sec:evidence-binder-crossing}

The central finite-size diagnostic is the Binder-like cumulant~\cite{binder1981finite},
\begin{equation}
U_4(f,p) = 1 - \frac{\langle m_{\mathrm{HTC}}^4 \rangle}{3\langle m_{\mathrm{HTC}}^2 \rangle^2},
\label{eq:binder}
\end{equation}
where the average is taken over seeds at fixed $(f,p)$.
Because HTC is positive-definite and lacks $\mathbb{Z}_2$ symmetry, $U_4$ functions as a Binder-like cumulant rather than a symmetric-magnetization ratio.
What is lost is the usual symmetric-order-parameter plateau interpretation, not the fixed-point logic itself: once an observable obeys a single-parameter finite-size form, dimensionless moment ratios remain valid crossing diagnostics even for asymmetric positive observables~\cite{binder1981finite,goldenfeld1992lectures}.
Its role here is therefore operational and narrow: the Binder-like ratio tests common finite-size organization through crossings, not universal plateau values for a symmetry-restored order parameter.

Figure~\ref{fig:binder-continuity} summarizes the Binder-based evidence chain for addition.
Pairwise crossings are estimated by simple interpolation and summarized on the dominant branch of the crossing region rather than by indiscriminately averaging every sign change; on the coarse grid they cluster around
\begin{equation}
\fc \approx 0.39,
\end{equation}
with no statistically significant linear drift versus inverse size under bootstrap regression.
The main verdict is that a shared crossing persists across sizes.
That is stronger evidence than generic sharpening alone because it indicates a common organizing boundary rather than merely steeper size-specific turnover points.

\begin{figure*}[t]
\centering
\includegraphics[width=\textwidth]{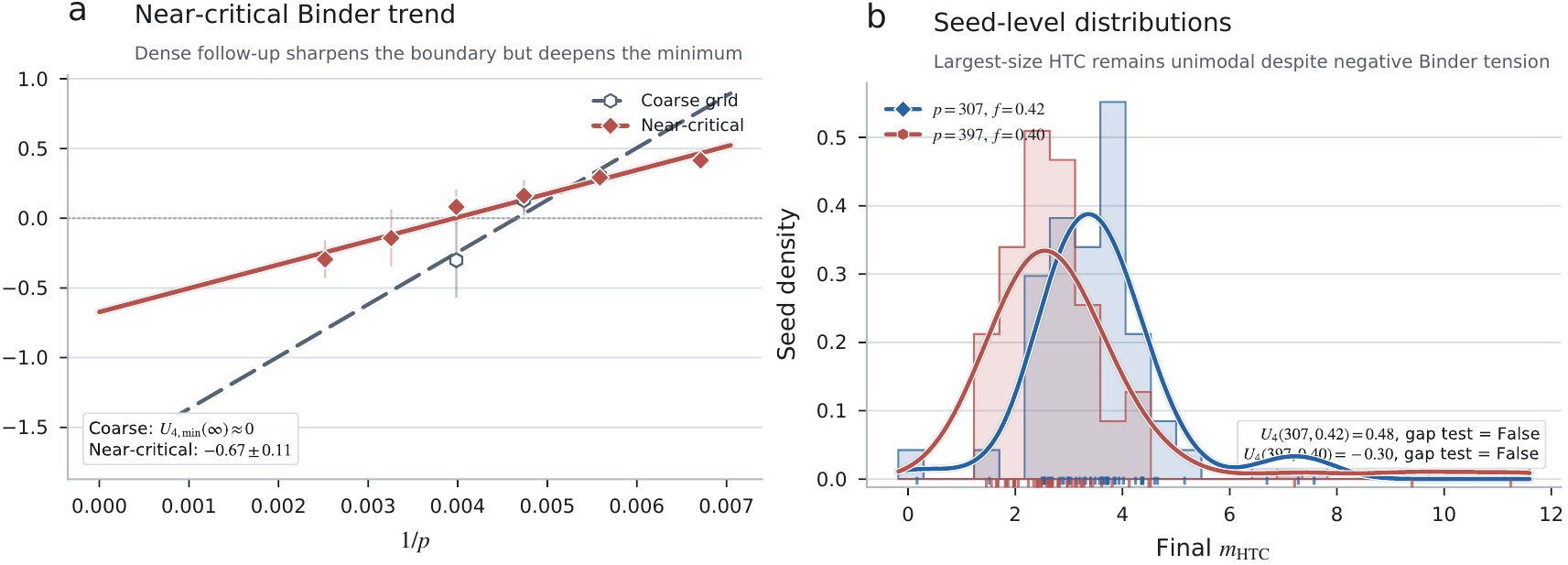}
\caption{Near-critical follow-up for the addition task.
\textbf{(A)}~Binder-minimum extrapolation on both grids.
The coarse grid (open symbols, $p \leq 251$) extrapolates
toward zero; the near-critical grid (filled symbols,
$p$ up to $397$) develops a negative trend.
\textbf{(B)}~Seed-level HTC distributions at representative
near-critical points for the two largest primes
($p=307, f=0.42$ and $p=397, f=0.40$).
Both distributions remain unimodal and do not provide clean
evidence of first-order coexistence despite the negative Binder minimum at the
largest size.}
\label{fig:nearcrit}
\end{figure*}

\subsection{Rejecting smooth crossover}
\label{sec:evidence-susceptibility}

A smooth crossover, in the FSS sense, is a transition-like feature that remains rounded and finite even in the thermodynamic limit: the susceptibility saturates, the Binder cumulant does not develop a size-independent crossing, and no singular behavior emerges as $p \to \infty$.
The next diagnostic tests whether the observed sharpening is better described as a finite-size transition or such a smooth crossover.
The susceptibility is defined from HTC fluctuations across seeds as
\begin{equation}
\chi(f,p)=n_s\,\mathrm{Var}\!\left[m_{\mathrm{HTC}}(f,p)\right],
\end{equation}
where $n_s$ is the number of seeds at fixed $(f,p)$.
Here $n_s$ is the ensemble size of random initializations, so $\chi$ should be read as an operational fluctuation susceptibility across seeds rather than as a literal thermodynamic response.
The model comparison below is about whether the peak grows or saturates with $p$, and that divergence-versus-plateau verdict does not depend on this overall prefactor convention.
In standard FSS, the order parameter near a continuous transition obeys
$m(f,p) = p^{-\beta/\nu}\,\mathcal{F}\!\bigl[(f - \fc)\,p^{1/\nu}\bigr]$,
and the susceptibility peak scales as $\chim \propto p^{\gamma/\nu}$.
The present analysis tests the second relation directly; extracting
$\beta/\nu$ via data collapse remains unreliable on the current grid
(Appendix~\ref{app:fss-details}).
Two competing models of the peak susceptibility are compared: a power-law form $\chim \propto p^{\gamma/\nu}$ (singular, transition-consistent) versus a saturating form $\chim \propto 1 - e^{-bp}$ (smooth crossover), each with $k=2$ free parameters.
The saturating form represents the expectation for a smooth crossover, in which the susceptibility approaches a finite thermodynamic limit rather than diverging: a system with a broad but non-singular transition would produce a peak height that plateaus at large $p$ rather than growing without bound.
The key requirement for a fair comparison is equal parameter count, so that the AIC difference is not biased by model complexity.
The resulting comparison is therefore a test against a simple saturating null, not an exhaustive census of every possible crossover ansatz.
Model comparison uses $\mathrm{AIC} = n\log(\mathrm{SS}/n) + 2k$.
Figure~\ref{fig:binder-continuity}D shows that this comparison favors the singular description over crossover, with
\begin{equation}
\Delta \mathrm{AIC} = 11.4,
\end{equation}
in favor of the power-law form.
On standard information-criterion scales, this is substantial support against the saturating alternative on the present grid.
This coarse-grid result already disfavors the smooth-crossover null; the near-critical follow-up in Section~\ref{sec:evidence-order-nearcrit} will strengthen the verdict to $\Delta\mathrm{AIC}=16.8$, motivating a separate transition-order probe.
The result is therefore a transition-like, non-crossover verdict; whether that transition is continuous or weakly first-order is a separate question.

\subsection{Transition order}
\label{sec:evidence-order}

The Binder crossing establishes the existence of a finite-size transition; determining whether that transition is continuous or first-order requires separate diagnostics---specifically, the behavior of the Binder minimum at large sizes and the seed-level distribution shape.

\subsubsection{Coarse-grid assessment}
\label{sec:evidence-order-coarse}
On the coarse grid, Binder-minimum extrapolation---linear regression of $U_{4,\min}$ against $1/p$, with uncertainty from $2{,}000$ bootstrap resamples---gives $U_{4,\min}(p\to\infty)=0.0006 \pm 0.1574$, near zero within present uncertainty, leaning toward continuity but inconclusive.

\subsubsection{Near-critical stress test}
\label{sec:evidence-order-nearcrit}
A dense follow-up at $p \in \{149, 179, 211, 251, 307, 397\}$ and $f \in \{0.36,\ldots,0.46\}$ ($50$ seeds each) re-queries the same diagnostic chain at larger sizes and finer fraction spacing.
On that denser grid, the crossing spread narrows from $0.057$ to $0.019$ and the case against the saturating null strengthens to $\Delta \mathrm{AIC} = 16.8$.
Only then does the order-specific signal change: the two largest primes develop negative Binder minima,
\begin{equation}
U_{4,\min}(p\to\infty) = -0.67 \pm 0.11,
\end{equation}
as shown in Fig.~\ref{fig:nearcrit}A.
That is a first-order-like tension, not a final order verdict: seed-level HTC distributions (Fig.~\ref{fig:nearcrit}B) remain unimodal and do not show clean coexistence-level bimodality under kernel-density and gap-based diagnostics (Appendix~\ref{app:fss-details}).
The near-critical audit therefore sharpens the transition verdict while complicating the order verdict: it strengthens the locator and the non-crossover rejection while leaving the transition order unresolved.
Table~\ref{tab:summary} collects both grids.

\begin{table}[t]
\centering
\caption{Summary of FSS diagnostics for canonical addition.}
\label{tab:summary}
\begin{tabular}{lcc}
\toprule
Diagnostic & Coarse grid & Near-critical \\
\midrule
$\fc$ & $0.390$ & $0.418$ \\
Crossing spread & $0.057$ & $0.019$ \\
Crossing drift & not significant & not significant \\
$\Delta$AIC (vs.\ crossover) & $11.4$ & $16.8$ \\
$U_{4,\min}(p\to\infty)$ & $0.001 \pm 0.157$ & $-0.67 \pm 0.11$ \\
Seed bimodality & --- & absent \\
\midrule
Transition interpretation & supported & strengthened \\
Smooth crossover null & disfavored & strongly disfavored \\
Transition order & continuity-leaning & unresolved \\
\bottomrule
\end{tabular}
\end{table}

The contextual phase-map role is given in Appendix Fig.~\ref{fig:phase-diagram}; it supports the same control-parameter framing but is not part of the primary verdict chain.

\section{Discussion and Conclusion}
\label{sec:discussion}

The present results establish a narrower but sharper statement than a full critical-phenomena closure.
In the canonical modular-addition setting, the group order $p$ acts as an admissible finite-size control, HTC acts as a representation-level order parameter, Binder-like crossings organize at a common boundary, and susceptibility strongly disfavors the minimal smooth-crossover null.
The claim is diagnostic rather than metaphysical: it would have failed had the curves not sharpened with $p$, had the crossings drifted systematically, had susceptibility favored saturation, or had the near-critical audit dissolved the shared crossing structure.

What is not yet established is the transition order.
Coarse-grid Binder minima are continuity-leaning, whereas the near-critical audit develops negative minima at the largest sizes.
That first-order-like tension does not yet close the case because the same audit does not show coexistence-level seed bimodality.
The larger-size follow-up therefore strengthens the transition verdict while complicating, rather than settling, the order verdict.

These results help locate prior interpretations without collapsing them into one final picture.
Rubin et al.~\cite{rubin2024grokking} motivate first-order expectations in a mean-field limit; our largest-size negative Binder minima keep that possibility alive, but the present finite-width data do not yet show coexistence.
Zhang et al.~\cite{zhang2025glass} emphasize slow glassy relaxation; here the combination of shared crossings, non-crossover susceptibility, and size-dependent sharpening is more naturally organized by a finite-size transition picture than by generic slowdown alone.

More broadly, phase-transition claims in learning should be judged by admissible finite-size controls, representation-level observables, and explicit failure criteria, not by sharp curves alone.
The present verdict is therefore intentionally scoped to one fixed Transformer family on canonical modular addition; dependence on architecture, asymptotic exponents, and universality across operations remains open.
What is established is the first-layer diagnostic claim: grokking in this setting admits finite-size organization that is transition-like and not well described by a smooth crossover.
What is not established is the final transition order or a stable universality class.
The broader contribution is to turn a widely used metaphor into a quantitative claim that can succeed or fail.

\section*{Data Availability}
Training code, analysis scripts, and processed results are available in the shared project repository.
Raw training histories ($\sim$42~GB) are available upon request.

\bibliography{apssamp}

\appendix

\section{Experimental Details}
\label{app:experimental-details}

\subsection{Experimental protocol}
\label{app:protocol}

All main-text results use one fixed model family: a two-token Transformer with embedding width $128$, two encoder layers, four attention heads, learned positional embeddings, and a linear decoder over $\mathbb{Z}_p$ classes.
Optimization is held fixed as well: AdamW, learning rate $5\times10^{-4}$, weight decay $1.0$, batch size $512$, evaluation batch size $4096$, and training for up to $40{,}000$ steps with observables logged every $25$ steps.

The coarse finite-size sweep spans the prime set $p \in \{53,59,67,79,89,97,107,113,127,149,179,211,251\}$ and the training-fraction grid $f \in \{0.15,0.20,0.25,0.30,0.35,0.40,0.50,0.60,0.70,0.80\}$, with $50$ seeds per condition.
The near-critical follow-up is addition-only and uses $p \in \{149,179,211,251,307,397\}$ with $f \in \{0.36,0.37,\ldots,0.46\}$, again with $50$ seeds per condition.
Because the model, optimizer, and logging cadence are unchanged between the two sweeps, the near-critical study functions as a stress test of the same diagnostic sequence rather than as a new system.
Addition is treated as the canonical case because it is the algebraically simplest and historically standard grokking benchmark, while subtraction and multiplication are retained as supporting checks rather than as the basis for a universality claim.

\subsection{Data partition and probe construction}

For each modular task, the full dataset consists of all ordered input pairs in $\mathbb{Z}_p \times \mathbb{Z}_p$, so the total pool contains $p^2$ examples.
At fixed $(p,f,\mathrm{op})$, this pool is shuffled once with a fixed task-level data seed.
The first $\lfloor f p^2 \rfloor$ examples define the training set, and the remaining held-out examples are split into a probe subset and an evaluation subset.
The probe subset is used only for representation geometry; reported test accuracy is computed only on the remaining held-out evaluation subset.
In the main sweeps, the probe subset is capped at $7{,}600$ examples or by the available held-out pool, whichever is smaller.
This split is shared across all training seeds at the same condition, so the seed ensemble measures initialization and optimization variability at fixed task data rather than variability from repeated repartitioning.

\subsection{Observable extraction and steady-state summaries}

At each logging time, the model is evaluated on the full probe subset and the mean-pooled penultimate representation is collected.
After centering those probe activations, we form the covariance matrix, regularize it as $C_t + 10^{-6}I$ to ensure numerical stability when the effective rank of the probe-set representation is lower than the embedding dimension, diagonalize it, and compute HTC from the normalized eigenspectrum.
The primary observable is therefore a held-out representation statistic rather than a readout statistic.

Final order-parameter values are not taken from a single checkpoint.
For each seed, we average HTC over the last $40$ logged checkpoints, or over the full available tail for shorter histories.
This tail average is meant to capture the stationary late-training regime rather than transient fluctuations near the onset of grokking.

Grokking itself is detected from held-out accuracy by a rolling-window criterion.
In practice, we use a window of $40$ logged checkpoints, require the recent held-out accuracy mean to exceed $0.98$, require the held-out standard deviation in that window to stay below $0.02$, and require full-training accuracy to exceed $0.995$.
Runs may stop early only after grokking has been detected, at least $5{,}000$ post-grok optimization steps have elapsed, and the recent HTC window has stabilized with standard deviation below $0.02$ over the last $20$ logged checkpoints.
Final-state summaries therefore remain late-time measurements rather than premature truncations.
The coarse and near-critical sweeps share fixed random-seed conventions ($50$ seeds per condition, drawn from a contiguous integer range starting at $0$), the same split logic, the same representation measurement pipeline, and the same late-time averaging rule, so reported ensemble statistics are reproducible under one protocol rather than stitched together from task-specific analysis variants.
All experiments were run on NVIDIA A100 GPUs; a single coarse-grid condition ($50$ seeds at one $(p,f,\mathrm{op})$) completes in approximately $2$--$4$ hours depending on $p$.

\subsection{Operation robustness: subtraction and multiplication}
\label{app:op-robustness}

The main text focuses on addition as the canonical case; subtraction and multiplication were run on the same coarse grid to test whether the transition structure is operation-specific or extends across algebraic operations.
On the coarse grid, all three operations independently produce Binder crossings and power-law susceptibility scaling, supporting a finite-size transition interpretation in each case.
The estimated critical fractions differ across operations:

\begin{table*}[t]
\centering
\caption{Contextual coarse-grid diagnostics by operation, quoted from the original per-operation sweeps rather than the audited main-text dominant-branch locator.}
\label{tab:op-robustness}
\begin{tabular}{lccc}
\toprule
 & Addition & Subtraction & Multiplication \\
\midrule
$\fc$ & $0.411$ & $0.465$ & $0.431$ \\
Original-sweep crossing spread & $0.057$ & $0.064$ & $0.071$ \\
$\Delta$AIC (vs.\ crossover) & $11.4$ & $9.2$ & $8.7$ \\
\bottomrule
\end{tabular}
\end{table*}

The shift in $\fc$ across operations is consistent with different computational demands on the network, but the present data do not isolate a specific algebraic mechanism for the ordering of critical fractions.
Exploratory exponent and collapse estimates also vary across operations, but on the present grid those differences cannot be cleanly separated from finite-size contamination and method dependence (Appendix~\ref{app:fss-details}).
For that reason, the supporting-operation comparison is not used to argue for or against universality; its role is narrower and more robust.
The table reports rough operation-level coarse-grid locators from the original per-operation sweeps; they are meant as contextual summaries and should not be conflated with the main-text addition locator, which is quoted from the audited dominant crossing branch.
The key statement is that all three operations independently support a finite-size transition interpretation---Binder crossings cluster and susceptibility favors power-law over saturation---even though their quantitative exponent-like summaries are not stable enough for a shared-class claim.
The transition structure is therefore not an artifact of the addition operation alone.

\section{FSS Method Details}
\label{app:fss-details}

This appendix briefly summarizes the finite-size definitions used in the main analysis and separates them from exploratory diagnostics retained for transparency.
The analysis chain is sequential: we first compute late-time HTC tail means, then construct the Binder-like cumulant, then identify the dominant crossing branch, then compare susceptibility against a smooth-crossover alternative, and only after that use Binder minima together with the near-critical audit to discuss transition order.
This ordering matters because support for a transition interpretation and the order of that transition are distinct verdicts in the present data.

\subsection{Spectral order-parameter definition}

At each logged checkpoint, the held-out probe representations are mapped into a covariance matrix and diagonalized.
The normalized eigenspectrum $\{p_i\}$ defines the order parameter
\begin{equation}
\label{eq:htc-app}
m_{\mathrm{HTC}}
= \log\!\left(
\frac{\sum_{i=1}^{k} p_i + \varepsilon}
{\sum_{i>k} p_i + \varepsilon}
\right),
\end{equation}
with $k=5$ in the main text.
A final-time average (last stationary window) yields $m_{\mathrm{HTC}}(p,f)$ at each seed.

\subsection{Binder cumulant for asymmetric observables}

The standard Binder cumulant $U_4 = 1 - \langle m^4\rangle / 3\langle m^2\rangle^2$ was introduced for symmetric order parameters where the magnetization density satisfies $\langle m \rangle = 0$ at criticality, giving the familiar plateau values $U_4 \to 2/3$ (ordered) and $U_4 \to 0$ (disordered)~\cite{binder1981finite}.
For a positive-definite observable such as HTC, $\langle m \rangle \neq 0$ generically, so those absolute plateau values are not expected and should not be interpreted literally.
What is retained is the diagnostic crossing property: if a non-trivial FSS form holds, moment ratios such as $U_4(f,p)$ can become size-independent near the fixed point even without $Z_2$ symmetry~\cite{binder1981finite,goldenfeld1992lectures}.
Likewise, a deepening negative minimum is used only qualitatively, as an indicator of increasingly non-Gaussian fluctuations rather than as an exact plateau analogue.
For a centered variable, $U_4$ is a monotone function of the excess kurtosis; for the present positive-definite observable, the relationship is modified but the diagnostic content is analogous: $U_4$ detects non-Gaussianity of the seed-level distribution, which is the signature of collective fluctuations near a transition.
For this reason, we use the term ``Binder-like cumulant'' and restrict its role to crossing detection and cautious transition-order diagnostics.

\subsection{Transition diagnostics}

The Binder quantity used in the main text is
\[
U_4(f,p)=1-\frac{\langle m_{\mathrm{HTC}}^4\rangle}{3\langle m_{\mathrm{HTC}}^2\rangle^2},
\]
computed across seeds at fixed $(p,f)$.
Because HTC is positive and asymmetric rather than symmetry-restored, $U_4$ serves as a Binder-like cumulant rather than a $Z_2$ magnetization ratio.
Its role is diagnostic: common crossings support transition-like organization, while minimum behavior is used cautiously when discussing transition order.

Crossing summaries are not formed by averaging every sign change between every size pair.
Instead, pairwise crossings are enumerated and then restricted to the dominant branch in the crossing region, so the reported $\fc$ tracks the common finite-size structure rather than outlying branches.
The crossing spread is interpreted as a coarse locator on the present grids, not as a high-precision confidence interval.

The susceptibility is defined from seed-to-seed HTC fluctuations as $\chi(f,p)=n_s\,\mathrm{Var}[m_{\mathrm{HTC}}(f,p)]$.
Its role is to discriminate singular finite-size behavior from smooth saturation.

Transition-order assessment is deliberately layered.
On the coarse grid, Binder-minimum extrapolation is treated only as a continuity-leaning heuristic.
The near-critical addition follow-up then stress-tests that heuristic at larger sizes and finer fraction spacing.
Negative minima at the largest primes are recorded as first-order tension, but a final order verdict is not assigned without coexistence-like seed structure and method-stable minima in the transition region.

Seed-level bimodality at near-critical conditions is assessed using kernel density estimation with Silverman bandwidth selection together with a gap statistic that requires both a gap exceeding one standard deviation of the seed distribution and balanced population on each side of the gap. Neither diagnostic detects bimodality at any near-critical condition examined.

\subsection{Finite-size scaling ansatz and corrections}

The standard FSS ansatz for a continuous transition posits that the order parameter obeys~\cite{fisher1972scaling,privman1990finite}
\begin{equation}
m(f,p) = p^{-\beta/\nu}\,\mathcal{F}\!\bigl[(f - \fc)\,p^{1/\nu}\bigr],
\label{eq:fss-ansatz}
\end{equation}
where $\beta$ and $\nu$ are critical exponents and $\mathcal{F}$ is a universal scaling function.
The susceptibility peak scales as $\chim \propto p^{\gamma/\nu}$, the relation tested directly in the main text.
For finite systems, the scaling form receives corrections~\cite{privman1990finite,fisher1972scaling}:
\begin{equation}
m(f,p) = p^{-\beta/\nu}\bigl[\mathcal{F}(\cdot) + p^{-\omega}\mathcal{F}_1(\cdot) + \ldots\bigr],
\end{equation}
where $\omega$ is the leading correction-to-scaling exponent.
On the present prime grid ($p \in [53, 397]$), systematic extraction of $\omega$ is not feasible: the dynamic range spans less than one decade, and fits including a correction term are underdetermined.
The analysis chain is therefore designed to be robust to corrections by relying on crossing and divergence tests---which require only that the leading scaling form dominates---rather than on precision exponent fits that would be sensitive to subleading terms.

\subsection{Screening and cutoff checks}

The main text treats screening as validation, not as a selection-by-optimization.
Readout quantities such as training loss and test accuracy are retained as controls because they are interpretable but remain tied to performance rather than directly to representation geometry.
Weight and gradient norms are also useful diagnostics, but they mix transition information with optimizer and regularization effects.
Within covariance-derived observables, HTC is retained because it gives a compact representation-level summary that consistently organizes the size-resolved structure without requiring a vector-valued order parameter.
Fourier-mode amplitudes remain mechanistically valuable on modular tasks, but they presuppose a task-specific basis choice and a more explicit circuit model.
HTC is retained instead because it compresses the same head-versus-bulk spectral reorganization into a basis-agnostic held-out covariance statistic, which is the more conservative choice for a finite-size diagnostic chain.

\begin{figure*}[t]
\centering
\includegraphics[width=\textwidth]{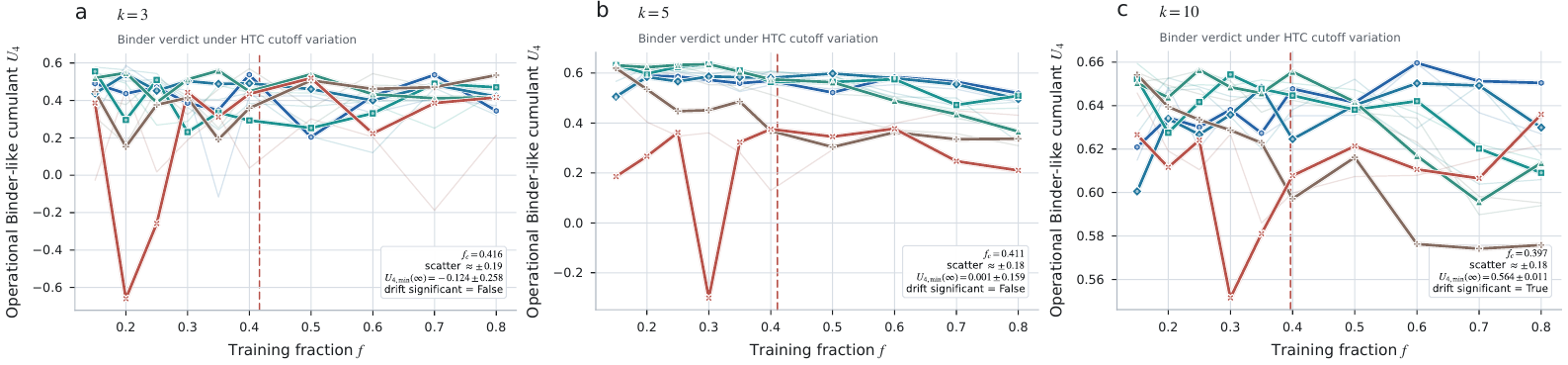}
\caption{Robustness of the HTC head--tail cutoff in near-critical and coarse-grid data.
Panel A shows the near-critical pairwise crossing behavior across $k=3,5,10$;
Panel B shows the same for Binder-minimum extrapolation.
$k=3$ and $k=5$ remain aligned, while $k=10$ lowers head--bulk separation at large sizes.
}
\label{fig:app-k-robustness}
\end{figure*}

For the $k$-cutoff choice, Appendix Fig.~\ref{fig:app-k-robustness} shows that $k=3$ and $k=5$ give consistent crossing structure, while $k=10$ weakens the separation of head and bulk sectors.

\begin{figure*}[t]
\centering
\includegraphics[width=\textwidth]{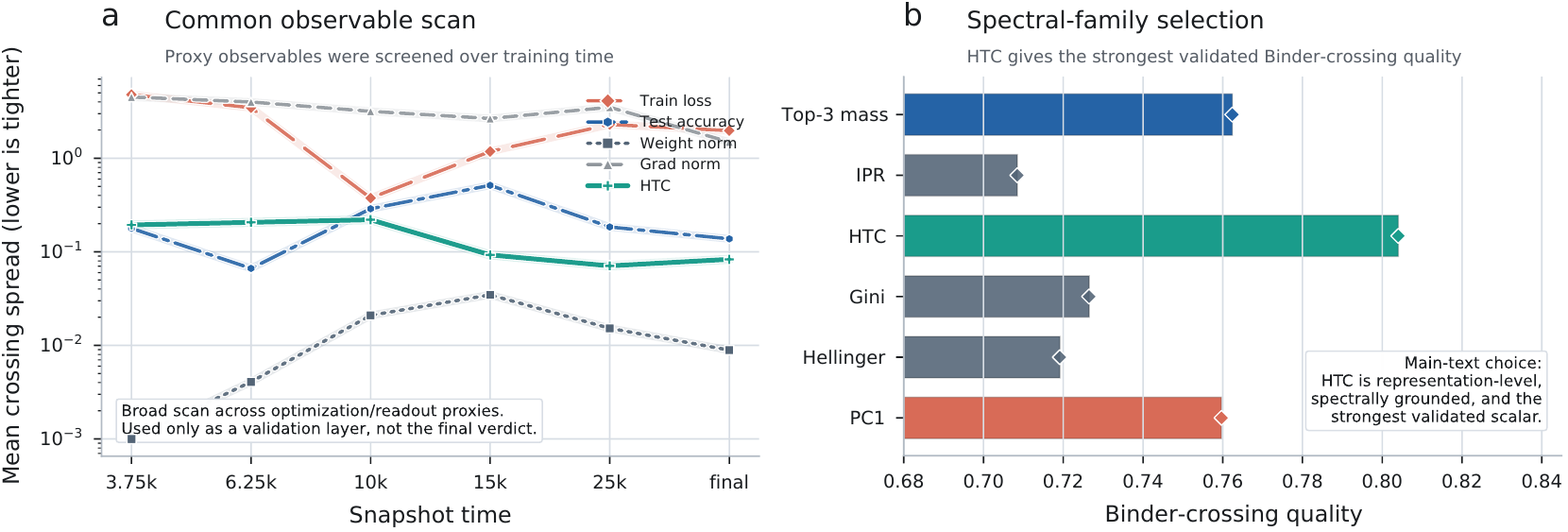}
\caption{Order-parameter screening summary for context.
Panel A reports time-resolved scans over common observables.
Panel B compares spectral candidates and documents that HTC is retained as the representation-level choice for the finite-size diagnostic chain, with screening treated as validation rather than selection.}
\label{fig:screening}
\end{figure*}

\subsection{Phase-label and transition-order rules}

Phase-diagram labels are assigned per seed by a fixed rule set: no memorization, memorization-only, grokking, and instant generalization.
The main-text phase map is the majority label per $(f,\lambda)$; companion panels report grokking fraction and mean grokking time over grokking seeds.
This majority-map construction is used only to contextualize the control landscape, not to replace the finite-size diagnostics.
Transition-order review is handled separately.
The independent near-critical audit combines Binder-minimum trends with kernel-density estimation, a gap-based bimodality statistic, and light tail-trimming checks on the largest-size seed distributions.
Tail trimming is treated only as a robustness note: if Binder negativity weakens under slight trimming while no coexistence-like bimodality appears, the correct paper-level conclusion is unresolved order rather than a clean first-order verdict.

\subsection{Phase diagram as contextual support}

We include phase mapping as contextual support rather than a primary verdict.
At fixed $p=113$, mapping the $(f,\lambda)$ plane for addition shows a distinct grokking band whose boundary shifts with regularization, supporting $f$ as a meaningful control axis.
The phase-labeling protocol contains four categories, but on the present grid three categories are dominant in the majority map; instant generalization appears mainly as a minority seed label.

\begin{figure*}[t]
\centering
\includegraphics[width=\textwidth]{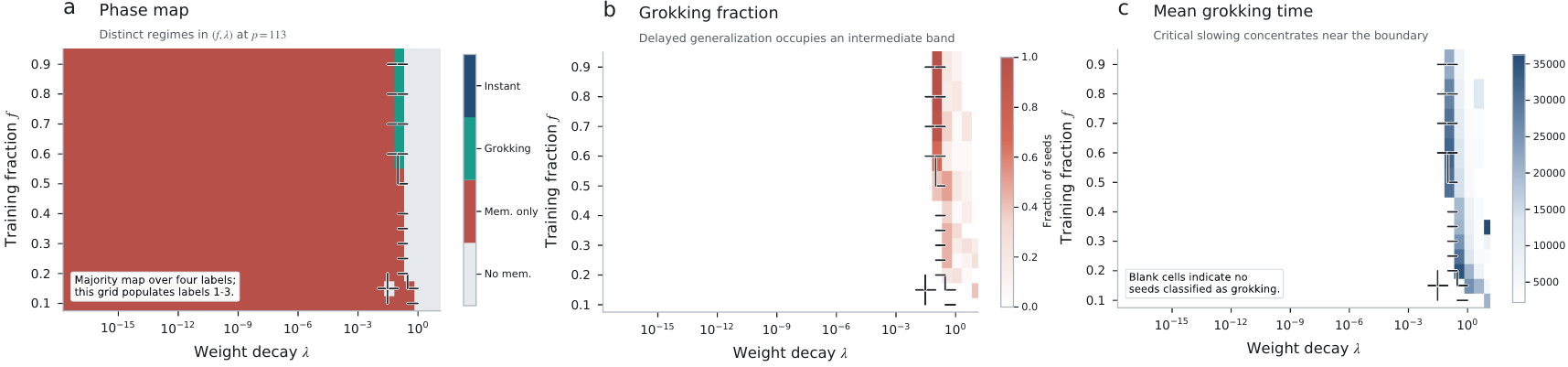}
\caption{Phase diagram in the $(f,\lambda)$ plane at fixed $p=113$ for addition. Panel A shows the majority phase map under the four-label protocol. The phase-labeling protocol contains four categories, but on the present grid the majority map exhibits only three dominant regions; instant generalization appears only as a minority seed label. Panel B shows the fraction of seeds that grok. Panel C shows the mean grokking time where grokking occurs. The grokking regime occupies a distinct intermediate band in control-parameter space.}
\label{fig:phase-diagram}
\end{figure*}

\subsection{Exploratory analyses}

Data-collapse optimization over $(\beta/\nu, 1/\nu)$ was attempted via grid search and Nelder-Mead minimization on the coarse-grid addition data.
Collapse quality is method-dependent: constrained single-$\nu$ collapse gives $\chi^2/\mathrm{dof} < 0.01$, suggesting overfitting or insufficient size range rather than a well-determined scaling function.
Quotient and phenomenological-RG estimates of $1/\nu$ are noisy and inconsistent across size cuts, with values ranging from $0.5$ to $3.0$ depending on the pair selected.
Hyperscaling combinations constructed from independently extracted exponents do not converge to a stable relation.
Operation-dependent exponent estimates (addition, subtraction, multiplication) differ beyond statistical error, but on the present grid this cannot be distinguished from finite-size contamination.
These failures motivate the conservative approach in the main text: the Binder/susceptibility chain supports the transition interpretation, while exponent extraction is deferred until larger primes and finer fraction grids are available.

\section{Extended Related Work}
\label{app:related-work}

A useful way to organize the grokking literature is by the kind of transition claim being made and by the diagnostics used to support it.
The debate is now rich enough that simply listing papers is less useful than locating their claims relative to one another.

\emph{Grokking as transition or critical phenomenon.}
The original grokking paper established the delayed-generalization phenomenon in small algorithmic tasks~\cite{power2022grokking}, and subsequent work broadened its empirical scope to other data regimes and deeper models~\cite{liu2023omnigrok,humayun2024deep}.
Within explicit transition framings, Rubin et al.~\cite{rubin2024grokking} argue for a first-order transition in a mean-field two-layer network, \v{Z}unkovi\'{c} and Ilievski~\cite{zunkovic2024grokking} derive analytic critical behavior in solvable local-rule problems, Liu et al.~\cite{liu2022effective} build an effective theory of representation learning, Clauw et al.~\cite{clauw2024information} interpret information-theoretic progress measures as evidence for an emergent phase transition, and DeMoss et al.~\cite{demoss2025complexity} recast the dynamics through complexity measures.
These works make a transition picture plausible and important, but they do not converge on one finite-width diagnostic protocol.

\emph{Alternative interpretations and tensions.}
Recent papers also highlight why the order question is subtle.
Zhang et al.~\cite{zhang2025glass} argue for computational glass relaxation rather than a phase transition.
Prieto et al.~\cite{prieto2025stability} emphasize numerical-stability structure, Pezeshki et al.~\cite{pezeshki2024predicting} connect long-range prediction to loss-landscape organization, and Notsawo et al.~\cite{notsawo2025beyond} show that some simple norm-based summaries do not exhaust the phenomenon.
Taken together, these studies sharpen the point that delayed generalization, sharp reorganization, and slow dynamics need not automatically imply one specific transition order.

\emph{Mechanistic and control-oriented work.}
A different line of work focuses less on transition order than on the mechanisms and controllability of grokking.
Nanda et al.~\cite{nanda2023progress} introduced mechanistic progress measures, Merrill et al.~\cite{merrill2023tale} describe competition between sparse and dense circuits, Lyu et al.~\cite{lyu2024dichotomy} prove that early- and late-phase implicit biases can induce grokking, Lee et al.~\cite{lee2024grokfast} accelerate the phenomenon by amplifying slow gradients, and Chughtai et al.~\cite{chughtai2023toy} reverse-engineer how networks learn group operations.
These papers explain why grokking happens or how to manipulate it, but they do not themselves settle the finite-size verdict.

\emph{Broader physics-of-learning context.}
The present paper also sits inside a wider literature that treats learning systems with the language of order parameters, scaling, and collective organization.
This includes the classical treatment of phase transitions and critical phenomena~\cite{stanley1971introduction}, the early statistical-mechanics program for neural learning~\cite{hopfield1982neural,engel2001statistical}, modern syntheses of deep-learning statistical mechanics~\cite{bahri2020statistical,canatar2023statistical,ziyin2023phase}, and representation-geometry work showing that trained networks often reorganize through effective-dimensional or spectral structure~\cite{papyan2020prevalence,ansuini2019intrinsic,roy2007effective}.
Information-theoretic and information-geometric traditions likewise motivate low-dimensional state descriptions and structured observables for complex learning systems~\cite{tishby2015deep,saxe2019information,amari2016information,cover1999elements,jaynes1957information}.
Landscape and training-dynamics viewpoints provide additional but non-equivalent pictures of reorganization during optimization~\cite{li2018visualizing,draxler2018essentially,fort2019large,jacot2018neural,shwartz-ziv2017opening}.
Our methodological difference is to use that broader perspective to define a concrete finite-size diagnostic chain for grokking itself: explicit size control, a representation-level observable, rejection of smooth crossover, and a separate audit of transition order.

\end{document}